\documentclass[letterpaper, 10 pt, conference]{ieeeconf}  

\IEEEoverridecommandlockouts                              

\usepackage{amsmath} 
\usepackage{amssymb}  

\usepackage{xcolor}
\usepackage{wrapfig}
\usepackage[export]{adjustbox}
\usepackage{url}
\usepackage{hyperref}
\usepackage{subcaption}

\usepackage{bm,bbm}

\title{\LARGE \bf
Towards Reinforcement Learning for Space Robotics:\\Fast Training and Reliable Real-World Transfer}

\author{Abhishek Naik$^{1}$, Michael Wu$^{2*}$, Michael O'Sullivan$^{3*}$, Colin Bellinger$^{3,4}$, Yunli Wang$^{1}$\vspace{-1mm}\\
\vspace{-0.5mm}
{\small \texttt{\{abhishek.naik,yunli.wang\}@nrc-cnrc.gc.ca, colin.bellinger@uottawa.ca}} \\
\vspace{-2mm}{\small ${^1}$National Research Council Canada \quad ${^2}$University of Waterloo \quad ${^3}$University of Ottawa \quad ${^4}$Vector Institute for Artificial Intelligence}
\thanks{*Equal contribution while interning at National Research Council Canada.}
}

\begin{document}

\maketitle
\thispagestyle{empty}
\pagestyle{empty}


\begin{abstract}
This work demonstrates the use of a fast simulation-based training framework to learn control policies using reinforcement learning. 
The hardware requirements are minimal and the simulation-learned policies transfer reliably to the real world. 
With this setup, we show that a quadruped robot can learn to traverse diverse real-world terrains reliably---including in a lunar-analog facility---after less than 20 minutes of training on a single consumer-grade GPU.
These results provide a glimpse of a future where reinforcement learning is a crucial tool in developing more capable and robust space robotic systems. 
Video demonstrations are available \href{https://youtu.be/vUFJiZdJaQ0}{here}.
\end{abstract}


\vspace{3mm}
The time is increasingly ripe to explore reinforcement learning (RL) for space applications. 
To this extent, there is growing interest in applying RL to orbital, planetary, and free-flying robotic systems \cite{APIARY_RL_Astrobee,RL-AVIST,RL_space_survey}. 
At the same time, two developments have fundamentally improved the applicability of reinforcement learning to robotics. 

First, advances in hardware-accelerated simulation and JAX-based learning frameworks now enable robot-control policies to be trained using thousands of parallel simulations running directly on a single GPU \cite{jax2018github, Zakka_Playground, IsaacSIM}. 
As a result, policies that previously required days of training can now be learned in minutes using consumer-grade hardware, enabling rapid prototyping and shorter scientific iterations. 
Second, sim-to-real techniques such as domain randomization (DR) are making simulation-learned policies increasingly reliable when transferred to physical robots. 
When combined, these advances enable extensive exposure to variations in simulated environmental conditions, sensor characteristics, and system dynamics, producing more robust real-world controllers while drastically reducing wall-clock training time.

We believe these advances make this an opportune time for the space-robotics community to research and develop RL as a practical tool for autonomous systems. 
In particular, the ability to learn behaviors rapidly in simulation and transfer them safely to hardware is well aligned with existing space-engineering workflows, which already rely heavily on digital simulation for system design, testing, and validation. 

In this paper, we highlight these developments through a series of case studies involving a quadruped robot. 
We first provide a brief overview of the simulation and learning infrastructure that enables rapid training, along with the use of DR for robust sim-to-real transfer. 
We then demonstrate that quadruped locomotion policies can be trained in a simple simulation environment in less than 20 minutes on a single consumer-grade GPU and deployed successfully across a range of real-world terrains, including a lunar-analog facility.


\section{General Insights from a Quadruped Platform}

We use a quadruped robot throughout this work.  
The ability of legged robots to traverse uneven terrain, step over obstacles, and access environments that may be difficult for wheeled systems makes them a compelling platform for exploring the role of RL in space robotics \cite{Frey_2026}. 
We view legged robots as a complementary mobility platform to traditional wheeled rovers for future space missions.
More broadly, the results presented here illustrate how modern RL methods can enable fast training, efficient experimentation, and reliable deployment of generic robotic systems operating in challenging environments.

We use quadruped locomotion as a representative and challenging benchmark for evaluating rapid simulation-based training and sim-to-real transfer.
We formulate quadruped locomotion as a continuous-control RL problem. 
Based on the standard Markov Decision Process formulation \cite{RLtextbook}, an agent observes a partial state of the world and learns to select actions that maximize long-term performance.
Our Unitree Go2 Edu robot is a 12-DoF quadruped having proprioceptive and optical sensors and onboard computing capabilities. 

\begin{figure}[t]
  \centering
  \includegraphics[trim={11cm 2.5cm 0 1cm},clip,width=0.4\textwidth]{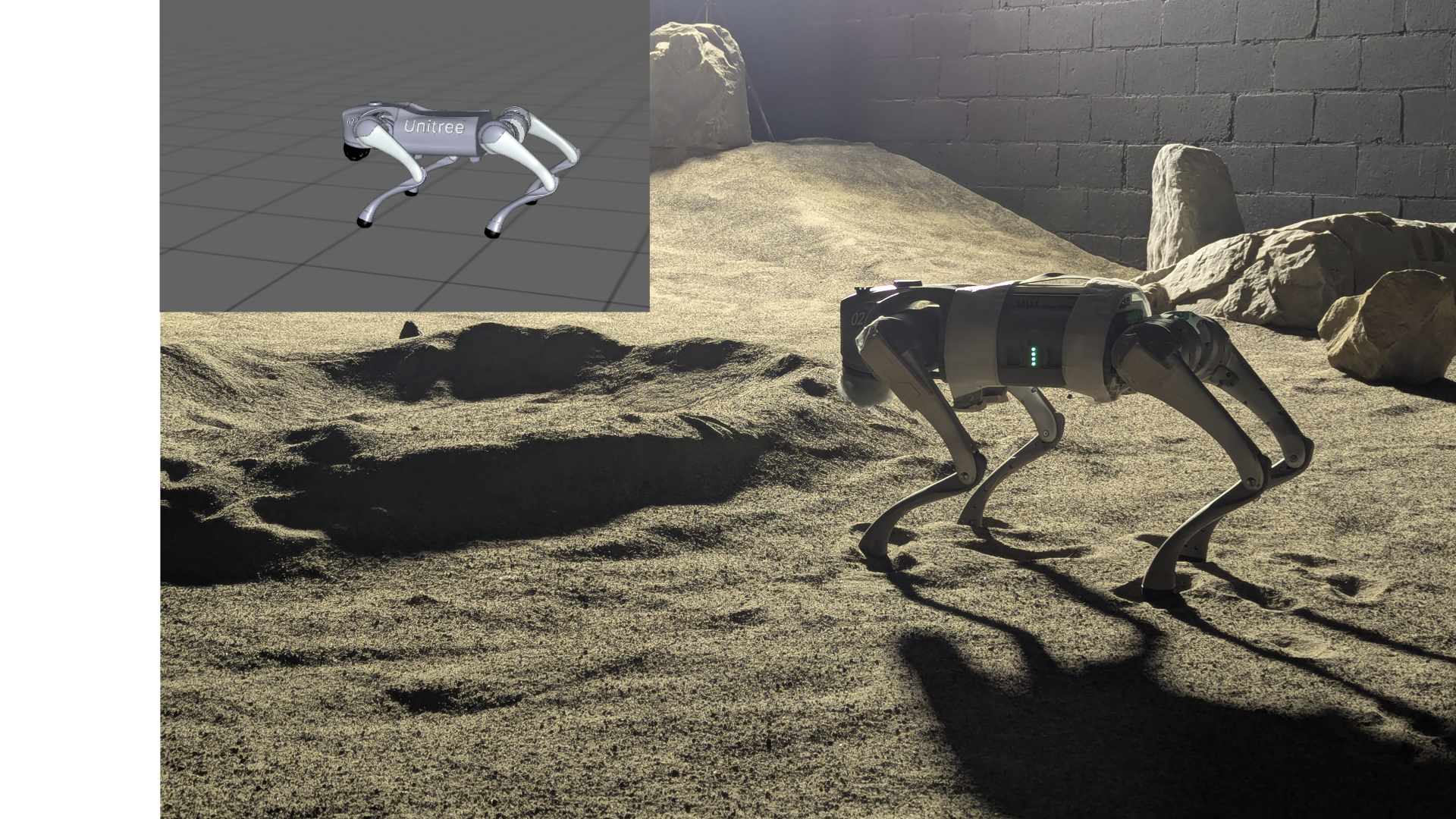}
  \caption{A photo of our quadruped robot walking in a lunar-analog facility after training with reinforcement learning in a simple simulation environment (inset).}
  \label{fig:rocky_moonyard}
  \vspace{-4mm}
\end{figure}


\section{Simulation Setup}
\label{sec:simulation_setup}

Two developments are particularly important for learning safe RL control policies with unprecedented speed: massively parallel simulation and GPU-native execution. 
Together, they enable complex control policies to be trained in minutes on consumer hardware.

\textbf{1. Parallel training paradigms}. Modern RL simulators can execute multiple robot environments in parallel \cite{Zakka_Playground, IsaacSIM}.
Algorithms (e.g., \cite{PPO}) can in turn process a batch of observations from the parallel environments and compute a batch of actions in parallel corresponding to each environment. 
This parallel paradigm dramatically increases data-collection throughput and can enable robust policies to be learned in minutes of wall-clock time rather than days. 

\textbf{2. GPU-native reinforcement learning}.
Large-scale machine-learning Python libraries such as JAX \cite{jax2018github} are enabling efficient use of hardware accelerators.
First, JAX-based Python code can be compiled rather than interpreted. 
Second, functions written in JAX can be run natively in parallel on a GPU.
When the simulation code, RL algorithms, and experiment pipelines are all written in JAX, the entire code can compiled into GPU-optimized code and run solely on a GPU---removing the significant overhead of CPU-GPU communication common in the traditional RL workflow. 
In our quadruped case studies presented below, we adapted the JAX-based robotic simulation environments and experiment pipeline within the Playground framework \cite{Zakka_Playground}. 

\begin{figure}[h]
    \centering
    \includegraphics[width=0.9\linewidth]{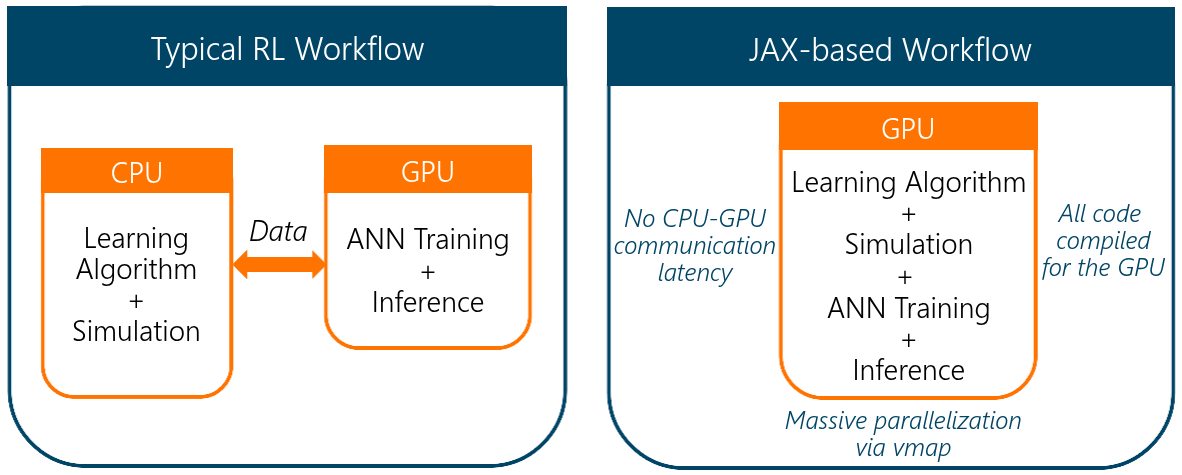}
    \caption{Comparison of the JAX-based workflow used in this paper (based on \cite{Zakka_Playground}) with the typical RL workflow.}
    \label{fig:JAX_workflow}
    \vspace{-1mm}
\end{figure}

\noindent \textbf{Implications for Space Robotics:} Many space-robotics workflows already rely heavily on simulation for design, testing, and validation. 
The same simulation-centric paradigm is increasingly enabling RL to train directly on large collections of digital scenarios before deployment. 
While not every space-robotics problem can be simulated well (e.g., applications involving non-rigid objects \cite{grapepicking}), the physics of many robotics tasks such as locomotion, navigation, manipulation, docking, and surface exploration can be simulated with sufficient fidelity. 
Consequently, RL with massively parallel simulation offers a pathway toward developing complex robotic behaviors for deployment in remote environments.


\section{Algorithms}

We employ PPO \cite{PPO}, a policy-gradient deep RL algorithm that updates a parameterized policy to maximize the expected sum of rewards while constraining the magnitude of policy updates, which improves training stability by preventing excessively large policy updates. 
PPO has emerged as the de facto approach for sim-to-real robotics \cite{47151, pmlr-v164-rudin22a}, in part because it is designed to benefit from the aforementioned parallel-environments training paradigm.
We used PPO for its accessibility; we expect the results in this paper to generalize to other RL algorithms that can be used in a parallel paradigm (such as SAC \cite{SAC}).

To improve sim-to-real transfer, we combined PPO with Domain Randomization (DR) \cite{DR}. 
During training, simulation parameters were varied across environments, including robot properties such as mass, and length of parts as well as terrain properties such as friction and slope. 
Training across a distribution of environments can reduce overfitting to a single simulator configuration; the resulting policies may be more robust to modeling errors and real-world variability. 
We validated that PPO and DR provide an efficient and practical recipe for learning robotic control policies that transfer reliably from simulation to the physical robot.


\section{Case Study 1: \\Zero-shot Walking on Flat Terrain}
\label{sec:friction}

This section focuses on walking on terrains with varying frictions. 
This is a fundamental task for earth-based and extraterrestrial mobile robots, who are likely to encounter a variety of surfaces, such as concrete, sand, gravel, puddles, ice, and so on, which they must navigate safely and reliably.

\textbf{Environment.} We created a simulation environment in which the robot is initialized in a flat open world (Fig.~\ref{fig:rocky_moonyard}, inset) and has to learn to follow joystick commands. 
There are three egocentric joystick inputs: linear velocity in the forward-backward direction, lateral velocity in the right-left direction, and rotational angular velocity about the vertical axis. 
The goal of the agent is to learn to move its limbs to achieve the joystick commands. 

The agent receives a 45-D observation vector including the angular velocity of the base, the projected gravity vector, the angular positions of all joints, the angular velocities of all joints, as well as the joystick velocity command.\footnote{The walking policies in this paper are trained without any visual inputs.}
The agent outputs the angular positions of all 12 joints, which are translated to the corresponding torques via a PD controller. 
Modeled after \cite{Zakka_Playground}, the reward is a weighted combination of factors encouraging adherence to the commanded velocities while penalizing slips, falls, and high magnitude of motor torques (see \cite{Zakka_Playground}'s Appendix B). 
Presently, the task is modeled as an episodic problem that terminates and resets every 1000 steps or when the robot falls---whichever comes first. 

We trained the policies using the Playground setup described in Section \ref{sec:simulation_setup} with a single agent interacting with 8192 parallel environments for a total of 250 million steps, which took around 11 minutes on an \textit{NVIDIA GeForce RTX 4070 Ti} GPU ($\sim$3 to compile the code and $\sim$8 to train). 

\textbf{Experiment.} We trained two policies with domain randomization over basic robot properties such as mass of joints, length of joints, and location of center-of-mass: each episode was initialized with a configuration sampled uniformly randomly around the default Playground values.
One of the policies was trained with an additional domain randomization over the sliding-friction coefficient between the robot's feet and the surface in the range $[0.6, 1.0]$; the baseline policy was trained with a fixed friction coefficient of $1.0$. 
We refer to these as \textit{FrictionDR} and \textit{BaseDR} policies respectively.
Our hypothesis was that the \textit{FrictionDR} policy 
would lead to more robust walking on different surfaces.

The two simulation-learned policies were transferred to the physical robot using \cite{Xu_URM} and evaluated in high and low frictions scenarios: a carpet and a wet vinyl sheet to simulate ice-like conditions. 
The evaluation involved 10 repeated trials that required the policy to execute a fixed sequence of joystick commands. 
We alternated the trials of the two policies to reduce unintended biases arising from evolving surface conditions. 

\begin{figure}[t]
    \centering
    \vspace{-3mm}
    \includegraphics[width=\linewidth]{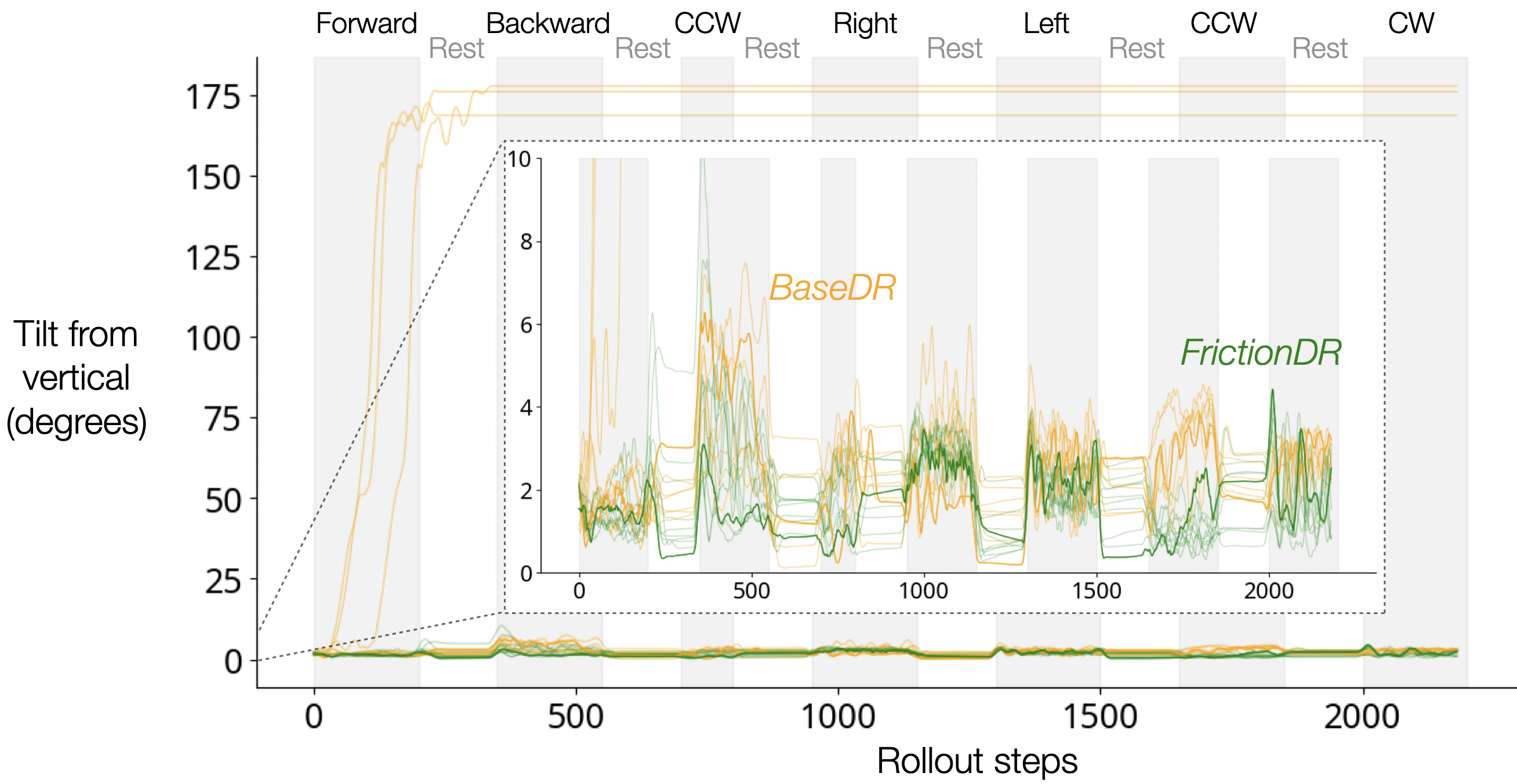}
    \caption{Evaluation of the stability of the simulation-learned policies with and without friction-based domain randomization on a real-world slippery surface.}
    \label{fig:friction_results}
    \vspace{-4mm}
\end{figure}

\textbf{Results}. Figure \ref{fig:friction_results} shows the results of real-world tests on the smooth low-friction surface. 
The x-axis denotes the steps over time and the y-axis denotes the tilt in the robot's torso w.r.t.~the upright direction, averaged over a moving window of 20 steps to smoothen out the variability in the data.\footnote{The tilt is one of the seven safety-related metrics that we tracked, including slips/falls, adherence to joint and torque limits, etc.}
The thin faded lines denote individual trials; the thick bold lines denote the median trial (fifth out of 10 trials based on the area under the curve). 
The gray areas denote periods with a specific non-zero command (specified at the top in black) while the white areas denote periods of rest.

The robot slipped and fell over in three out of the 10 trials of the \textit{BaseDR} policy when walking forward; it did not fall with the \textit{FrictionDR} policy in any of the 10 trials.
The robot slipped with both policies when walking backward.
Thereafter, there were minor differences in the performance of the two policies.
The behavior is quantified in Figure \ref{fig:friction_results}. 
In the high-friction carpet-walking scenario, the two policies were relatively indistinguishable and the corresponding plot has been omitted due to space constraints. 

Overall, both the qualitative observations and the quantitative data indicated that the policy trained with additional friction-based DR in simulation resulted in more stable walking behavior in the real world.
In further outdoor tests of the \textit{FrictionDR} policy, the robot traversed all the tested surfaces---concrete, asphalt, pebbles, grass, sand---reliably, without falling (see the accompanying video).


\section{Case Study 2: \\Zero-shot Walking on Uneven Terrain}
\label{sec:moonyard}

Many real-world applications involve traversing uneven terrain (not necessarily smooth and/or flat). 

\textbf{Environment}. The only modification we made to the simulation environment was to make the surface inclined. 
The agent's goal, observations, actions, and the reward signal remained the same. 
This was because a guiding principle of this work was to minimize the simulation-design effort and test the agent's ability to learn a robust policy in large-scale simulations for zero-shot transfer.

\textbf{Experiment}. 
We trained a policy with a curriculum of domain randomization over increasing degrees of slope: first 200M steps at zero slope, then 50M with slopes among \{0, 5, 10\} degrees, then 50M with \{0, 5, 10, 15\}, finally 50M more steps with \{0, 5, 10, 15, 20\}. 
Training a policy under an automated curriculum of this kind would take about $15$ minutes on our single GPU.
This new \textit{SlopesDR} policy was also transferred to the robot and tested alongside two policies: the \textit{FrictionDR} policy from the previous experiment and a new \textit{NoDR} policy trained without \textit{any} domain randomization.

\begin{figure}[b]
    \centering
    \vspace{-3mm}
    \includegraphics[trim={1cm 4cm 1cm 2cm},clip,width=0.9\linewidth]{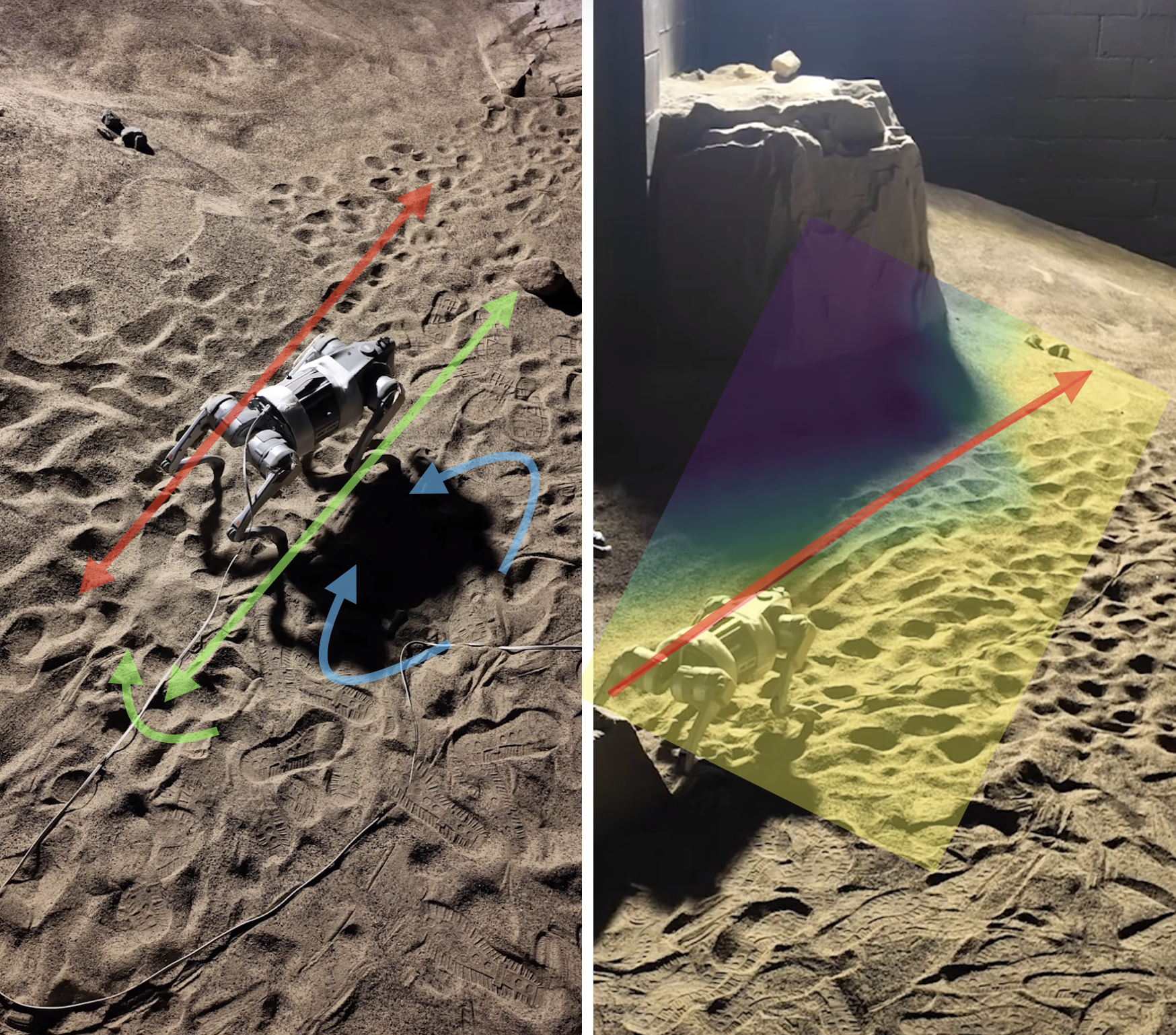}
    \caption{The two scenarios tested at the lunar-analog facility. \textit{Left:} Walking on a flat surface. \textit{Right:} Walking across the side of an incline (the inset heatmap denotes the inclination---yellow is flat; blue is steep). The arrows indicate the movements tested.}
    \label{fig:moonyard_scenarios}
    \vspace{-4mm}
\end{figure}

We deployed these three policies within a lunar-analog facility. 
We tested two scenarios: walking on a flat surface, and walking across the side of an incline (see Figure \ref{fig:moonyard_scenarios}). 
The former involved the whole sequence of movements as before while the latter (harder) scenario only tested if the quadruped can walk forward across the side of the incline. 
As before, we randomly alternated the 10 trials of the three policies. 

\begin{figure}[t]
\centering
    \begin{subfigure}{0.99\linewidth}
      \centering
      \includegraphics[width=0.99\linewidth]{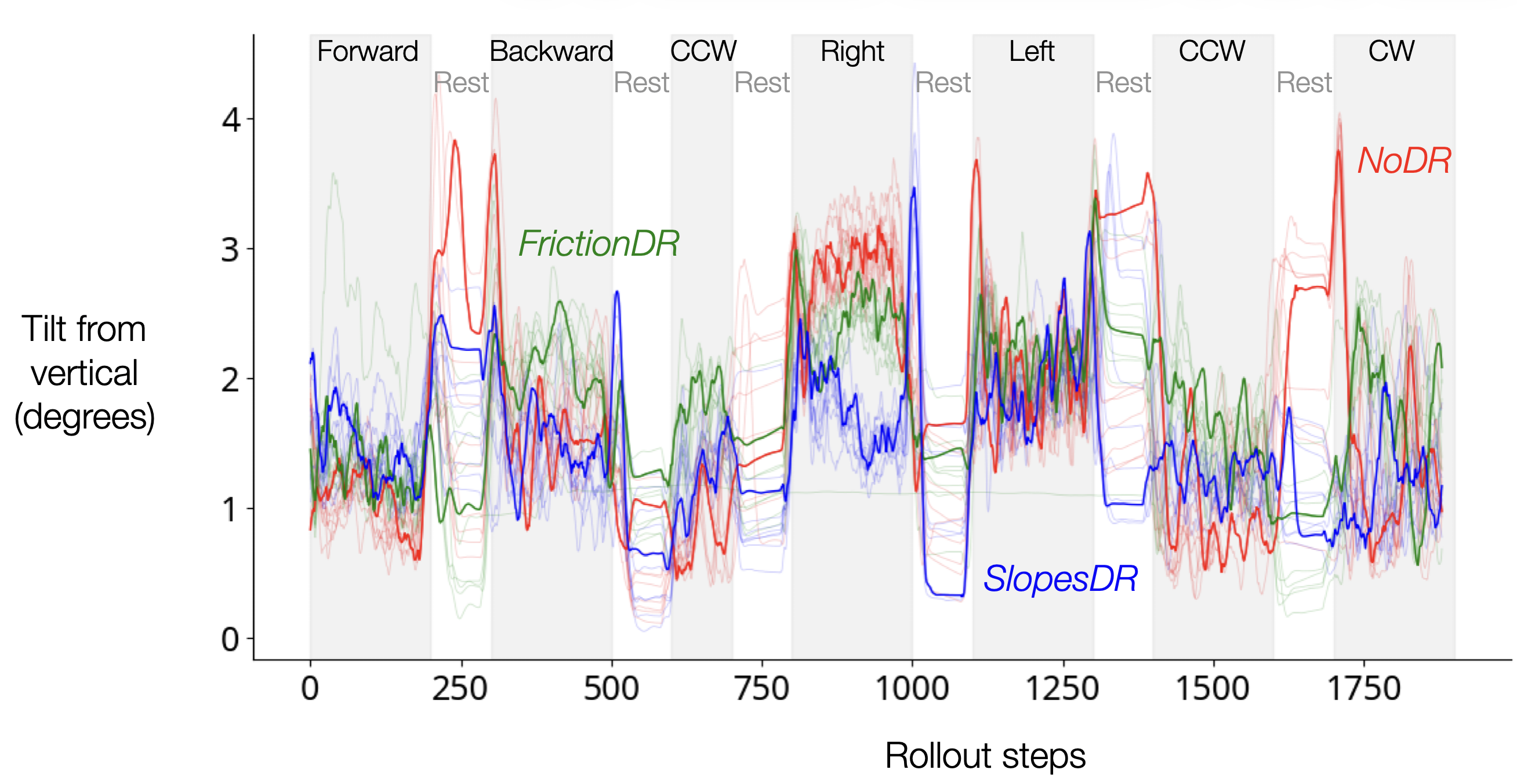}
    \end{subfigure}
    \begin{subfigure}{\linewidth}
      \centering
      \includegraphics[width=0.99\linewidth]{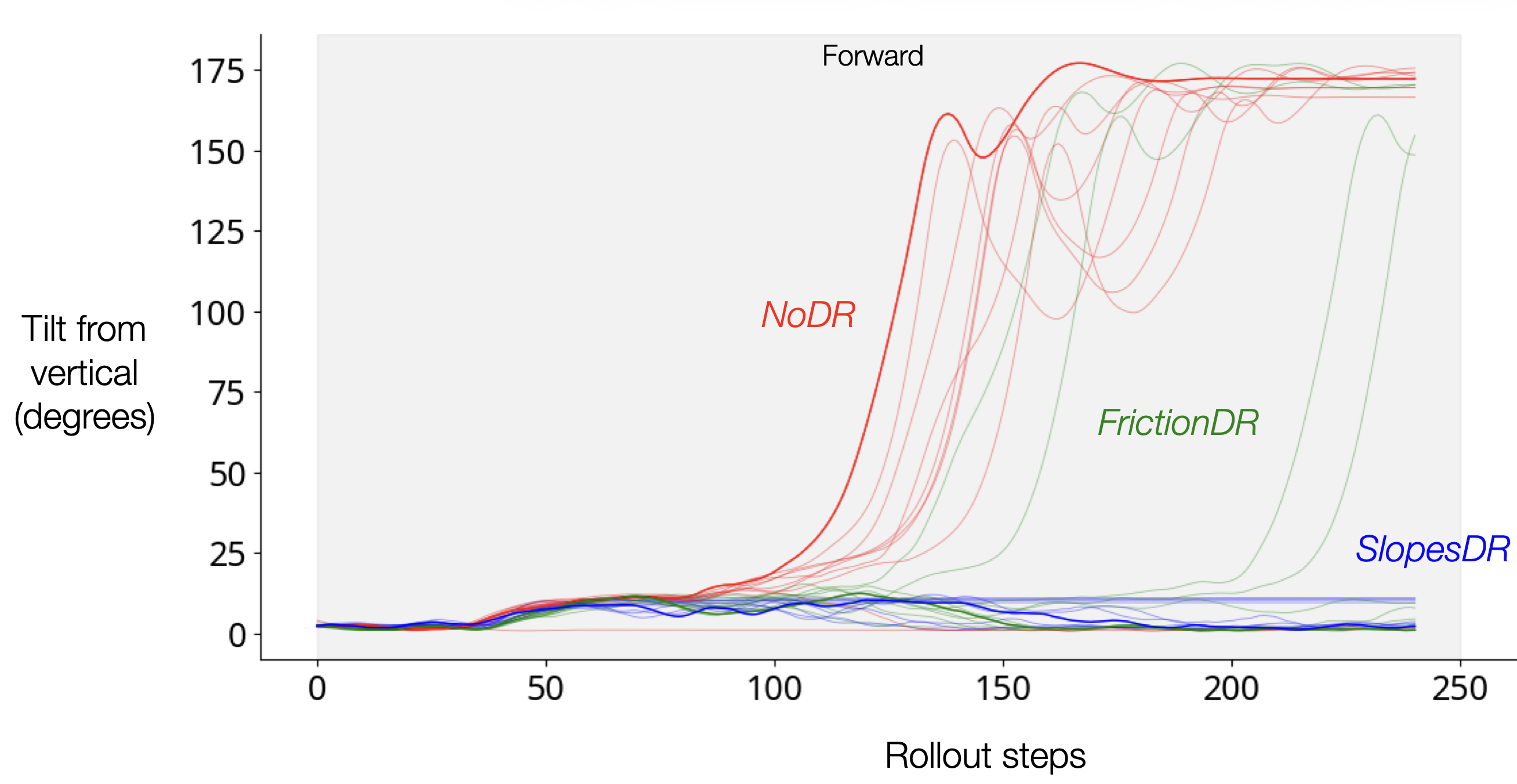}
    \end{subfigure}
\caption{Results of the real-world evaluations in the lunar-analog facility. \textit{Top:} Evaluations on a flat surface. \textit{Bottom:} Evaluations across the side of an incline.}
\label{fig:moonyard_results}
\vspace{-4mm}
\end{figure}

\textbf{Results}. Figure \ref{fig:moonyard_results} shows the results of the two experiments. 
The top row shows that the performance of the three policies was relatively similar in the flat scenario. 
The two policies trained with domain randomization were expected to do well in this test---which they did. 
Notably, the policy trained without any domain randomization also performed well. 

The differences between the policies were more apparent in the second scenario (bottom row of Figure \ref{fig:moonyard_results}). 
The robot fell over in nine of the 10 trials using the \textit{NoDR} policy within a couple of seconds of climbing the side of the incline. 
This is unsurprising given the \textit{NoDR} policy was only trained on flat ground in simulation. 
With the \textit{FrictionDR} policy, the robot fell in some trials and succeeded in others, which indicates that domain randomization in simulation over the friction and some robot attributes helped in generalizing slightly to the unseen inclined terrain.
The \textit{SlopesDR} policy either succeeded in walking across the side of the incline, or---interestingly---stopped midway on the side of the incline, resulting in two modes in the data towards the end of the trials: either the robot was back on flat ground with an upright torso or was stationary on the side of the incline with an inclined torso.
It is possible that the agent learned this behavior in simulation to avoid getting the penalty for falling when learning to walk on the different inclined surfaces.

Overall, domain randomization over slopes in simulation helped learn a policy that can traverse uneven terrain. 
The robot also successfully traversed inclined planks and grassy hills with the \textit{SlopesDR} policy in our other real-world tests (see the supplementary video).

\section{Conclusions and Future Work}
\label{sec:future_work}

We showcased recent advances in RL that enable learning complex control policies quickly in simulation that can also be reliably transferred zero-shot to the real world.
The massive speedup in training robust control policies enables rapid prototyping and short scientific iterations, which can result in faster research progress. 
We hope this work inspires the space-robotics community to consider reinforcement learning as part of the toolkit for building safe and effective robots. 

The particular demonstrations in this paper with a quadruped robot are the first of many steps towards adaptive robots that start with simulation-learned behavior and continue to learn in the real world.
The primary direction of future work is to build upon the low-level walking policies to achieve high-level goals in tasks involving construction, firefighting, agriculture, and so on (see the supplementary video for a proof-of-concept application of the robot autonomously using the low-level walking policy to search for and go towards an object of interest---a football).
There are many interesting research questions along the way, such as considering other efficient RL algorithms like SAC \cite{SAC} that can be run in parallel, testing continual onboard learning with algorithms like AVG \cite{AVG}, and considering other action spaces such as joint-velocity control or direct torque control that may be easier to learn with or result in richer behavior. 

An important direction of future work is to compare RL-based policies with classical/model-based controllers and develop ideas that extract the best of both kinds of approaches.


\addtolength{\textheight}{-12cm}   


\section*{Acknowledgments}

The Canadian AI Safety Institute supported a part of this work.
The lunar-analog facility is at Mission Control Space Services' office in Ottawa: \href{https://missioncontrolspace.com/about/facilities/moonyard}{missioncontrolspace.com/about/facilities/moonyard}. 
We also thank Nicolas S\'{e}guin and team for some of the photographs and video clips used in this paper and the supplementary video.



\bibliographystyle{IEEEtran}
\bibliography{bibliography}


\end{document}